\documentclass[twocolumn]{article}
\usepackage{amsmath,amsfonts}
\usepackage{algorithmic}
\usepackage{algorithm}
\usepackage{array}
\usepackage[caption=false,font=normalsize,labelfont=sf,textfont=sf]{subfig}
\usepackage{textcomp}
\usepackage{stfloats}
\usepackage{url}
\usepackage{verbatim}
\usepackage{graphicx}
\usepackage{cite}
\usepackage{float}
\usepackage{tikz}
\usepackage{makecell} 
\usepackage{longtable}

\usepackage{xcolor}
\definecolor{olive}{rgb}{0.33, 0.42, 0.18}
\usepackage[final]{microtype}

\usepackage{url}

\usepackage{color}
\usepackage{array}
\usepackage{tabularx}
\usepackage{hyperref}
\usepackage{listings}

\begin{document}
\renewcommand{\arraystretch}{1.3}

\twocolumn[\begin{@twocolumnfalse}

\title{
SVH-BD : Synthetic Vegetation Hyperspectral Benchmark Dataset for Emulation of Remote Sensing Images
}

\author{
Chedly Ben Azizi, Claire Guilloteau, Gilles Roussel, and Matthieu Puigt\thanks{The authors are with Univ. Littoral C\^ote d'Opale, LISIC -- UR 4491, 62219 Longuenesse, France. E-mail: firstname.lastname@univ-littoral.fr.}
}

\date{}

\maketitle

\abstract{
This dataset provides a large collection of 10\ 915 synthetic hyperspectral image cubes paired with pixel‑level vegetation trait maps, designed to support research in radiative transfer emulation, vegetation trait retrieval, and uncertainty quantification. Each hyperspectral cube contains 211 bands spanning 400–2500 nm at 10 nm resolution and a fixed spatial layout of $64 \times64$ pixels, offering continuous simulated surface reflectance spectra suitable for emulator development and machine‑learning tasks requiring high spectral detail. Vegetation traits were derived by inverting Sentinel‑2 Level‑2A\cite{sentinelhub2023} surface reflectance using a PROSAIL‑based lookup‑table approach\cite{jacquemoud1990prospect,verhoef1984light,jacquemoud2009}, followed by forward PROSAIL simulations to generate hyperspectral reflectance under physically consistent canopy and illumination conditions. The dataset covers four ecologically diverse regions—East Africa, Northern France, Eastern India, and Southern Spain—and includes 5th and 95th percentile uncertainty maps as well as Sentinel‑2 scene classification layers. This resource enables benchmarking of inversion methods, development of fast radiative transfer emulators, and studies of spectral–biophysical relationships under controlled yet realistic environmental variability.
}

\hfill

\end{@twocolumnfalse}]

\clearpage
\newpage

\section*{Specifications table}


\begin{tabular}{ p{.22\textwidth}  p{.70\textwidth} }

        \textbf{Subject} & \textbf{Earth \& Environmental Sciences} \\
        \hline
        Specific subject area & Hyperspectral Remote Sensing, Vegetation Trait Retrieval, Radiative Transfer Model Emulation \\
        Type of data & Reflectance hyperspectral images, Vegetation bio-optical parameter maps \\
        Data collection & The dataset was generated through a multi-step pipeline combining satellite data preprocessing, radiative transfer model inversion, and forward hyperspectral simulation. Sentinel-2 Level-2A \cite{sentinelhub2023} products were first selected across four geographic regions (East Africa, France, Spain, India). The Sentinel-2 tiles were acquired from Level-2A surface reflectance products via the Harmonised Data Access service, with acquisition dates between 1 and 6 May 2023. The relevant spectral bands were extracted from each Level-2A tile, specifically ['B1', 'B2', 'B3', 'B4', 'B5', 'B6', 'B7', 'B8', 'B8A', 'B9', 'B11', 'B12'], and were subsequently cropped and harmonized to obtain spatially consistent multispectral inputs of 64 by 64 pixels, and 20m ground sampling distance.  These inputs were then inverted using a PROSAIL-based look-up table inversion \cite{jacquemoud1990prospect,verhoef1984light,jacquemoud2009} to estimate pixel-level vegetation biophysical parameters. Finally, the retrieved parameters were used to drive a forward PROSAIL simulation to produce hyperspectral reflectance cubes in the range of 400-2500nm. \\
        Data source location & Institution: Laboratoire d’Informatique, d’Image et du Signal de la Côte d’Opale
                                \newline City/Town/Region: Longuenesse
                                \newline Country: France \\
        Data accessibility & Repository name: SVH-BD : Synthetic Vegetation Hyperspectral Benchmark Dataset for Emulation of Remote Sensing Images
        \newline Data identification number: \url{https://doi.org/10.5281/zenodo.18660571}
        \newline Direct URL to data: \url{https://zenodo.org/records/18660571} \\
        Related research article & None
        
\end{tabular}

\clearpage

\section*{Value of the data}
\begin{itemize}
    \item The dataset provides 10 915 synthetic hyperspectral image cubes with pixel‑level vegetation trait maps, offering a combination of dense spectral information and explicit vegetation trait annotations that are typically unavailable in observational remote sensing archives. Each hyperspectral cube includes 211 spectral bands from 400–2500 nm at 10 nm resolution, supplying continuous reflectance spectra suitable for emulator development and machine‑learning tasks requiring high spectral detail.
    \item The hyperspectral cubes have a  fixed spatial resolution of 64 x 64 pixels and a uniform 20 m ground sampling distance, which ensures spatial consistency across all scenes, enabling both pixel‑wise and spatial-wise learning under controlled and comparable conditions. 
    \item The forward and inverse modeling steps are guided by plant biophysical constraints, region‑specific parameter distributions and soil types, ensuring that simulated reflectance and retrieved traits reflect realistic environmental variability and remain consistent with the ecological characteristics of each region.
    \item The dataset includes uncertainty maps for vegetation trait retrievals, providing 5th and 95th percentile estimates that support the evaluation of model robustness, the study of uncertainty propagation, and the development of methods that explicitly account for retrieval variability. Additional Sentinel‑2 classification maps are provided for each scene, offering contextual information that supports land‑cover‑aware analysis, and enables studies that integrate spectral, biophysical, and categorical landscape information.
\end{itemize}
\section*{Background}

The increasing availability of multispectral and hyperspectral satellite missions has transformed remote sensing into a data-rich field capable of monitoring vegetation structure, function, and biochemical status at regional to global scales. Despite this progress, the development of data-driven retrieval and emulation models remains constrained by the limited availability of pixel-level vegetation trait annotations. Large remote sensing archives provide extensive observational coverage but rarely include physically consistent biophysical variables such as chlorophyll content, leaf water content, or leaf area index. This lack of reference data limits the training, benchmarking, and validation of emerging machine learning approaches aimed at linking canopy reflectance to underlying vegetation properties. This dataset has therefore been developed to fill this gap, enabling benchmarking of emulation models and supporting research in trait retrieval, uncertainty propagation, and physically informed machine learning. Physically based simulation has become an essential strategy to address this gap. Radiative transfer models such as PROSPECT \cite{jacquemoud1990prospect} and SAIL \cite{verhoef1984light} allow controlled generation of leaf and canopy reflectance by explicitly modeling the influence of biochemical composition, structural parameters, soil background, and viewing geometry. These simulations can produce synthetic datasets that are consistent with physical laws and tailored to the parameter ranges encountered in real data. Such a simulated dataset would serve multiple purposes, such as enabling model benchmarking, support the development of fast emulators that approximate radiative transfer models at significantly reduced computational cost, and facilitate large-scale inversion workflows for vegetation trait retrieval.
\section*{Data description}
The dataset contains $10\ 915$ hyperspectral image cubes, each with spatial dimensions of $64 \times 64$ pixels and 211 spectral bands. The spectral range spans 400–2500 nm with a fixed 10 nm sampling interval, and all images share a 20 m ground sampling distance (GSD). Each hyperspectral cube is paired with a trait file (\texttt{traits.tif}) containing 16 leaf‑, canopy‑, and observation‑level parameters. Additional files provide uncertainty estimates for trait retrievals (\texttt{p5.tif}, \texttt{p95.tif}) and Sentinel‑2 scene classification maps (\texttt{quality\_scene\_classification.img} with metadata in \texttt{quality\_scene\_classification.hdr}). Table~\ref{tab:content} provides a summary of the content of each tile folder. Details about the traits can be found in Table~\ref{tab:prosail_params}.

\begin{figure}[h!]
\centering
\begin{lstlisting}[basicstyle=\ttfamily\footnotesize]

dataset/
|-- <REGION_ID>/
|   |-- <TILE_ID>/
|   |   |-- surf_refl.tif
|   |   |-- traits.tif
|   |   |-- p5.tif
|   |   |-- p95.tif
|   |   |-- quality_scene_classification.img
|   |   `-- quality_scene_classification.hdr
|   `-- ...
`-- ...
\end{lstlisting}

\caption{Folder hierarchy of the dataset.}
\end{figure}

\begin{table*}[ht]
    \centering
    
    \caption{Summary of files provided for each tile}
    \begin{tabularx}{\textwidth} { >{\raggedright\arraybackslash}X >{\raggedright\arraybackslash}X }
        \textbf{Filename} & \textbf{Description} \\
        \hline
        \texttt{surf-refl.tif} & Hyperspectral surface reflectance cube ($64 \times 64 \times 211$) \\
        \texttt{traits.tif} & Retrieved bio-optical parameters (16 traits) \\
        \texttt{p5.tif} & 5th percentile uncertainty map for trait retrievals. \\
        \texttt{p95.tif} & 95th percentile uncertainty map for trait retrievals. \\     
        \texttt{quality\_scene\_classifcation.img} & Sentinel-2 scene classification map. \\
        \texttt{quality\_scene\_classifcation.hdr} & Metadata associated with the classification map.
    \end{tabularx}
    \label{tab:content}
\end{table*}

\begin{table*}[ht]
    \centering
    \caption{Parameter ranges for the combined leaf (PROSPECT-D) and canopy (4SAIL) RTM (PROSAIL).}
    \begin{tabular}{llll}
        
        \textbf{Symbol }& \textbf{Description} & \textbf{Unit} & \textbf{Range} \\
        \hline
        \multicolumn{4}{l}{Leaf Biochemical Parameters (PROSPECT\textendash D)} \\
        \hline
        $N$ & Leaf structure parameter & -- & $[1,\;2.5]$ \\
        $C_{ab}$ & Leaf chlorophyll a+b & $\mu g\,cm^{-2}$ & $[0,\;160]$  \\
        $C_{ar}$ & Leaf carotenoids & $\mu g\,cm^{-2}$ & $[0,\;60]$ \\
        $C_{ant}$ & Leaf anthocyanins content & $\mu g\,cm^{-2}$ & $[0,\;5]$ \\
        $C_{brown}$ & Brown pigments & -- & $[0,\;1]$ \\
        $C_{w}$ & Equivalent water thickness & $g\,cm^{-2}$ & $[0,\;0.07]$ \\
        $C_{m}$ & Dry matter content& $g\,cm^{-2}$ & $[0,\;0.1]$ \\
        
        \hline
        \multicolumn{4}{l}{Canopy Structural Parameters (SAIL)} \\
        \hline
        
        LAI & Leaf area index & -- & $[0,\;10]$ \\
        LIDFa & Average leaf angle & degrees & $[30,\;70]$ \\
        LIDFb & Bimodality & -- & $0$ \\
        TypeLIDF & Leaf inclination distribution function & -- & 1 \\
        h$_{spot}$ & Hotspot parameter & -- & $[0.01,\;0.5]$ \\
        
        \hline
        \multicolumn{4}{l}{Soil Parameters} \\
        \hline

        $\rho_{soil}$ & Soil spectrum & -- & Library( \cite{DVN2021})  \\
        
        \hline
        \multicolumn{4}{l}{Observation Geometry} \\
        \hline
        $\theta_s$ & Solar zenith & degrees & $[15,\;80]$ \\
        $\theta_v$ & View zenith & degrees & $[0,\;35]$ \\
        $\phi$ & Relative azimuth & degrees & $[100,\;150]$ \\
        \hline
    \end{tabular}
    \label{tab:prosail_params}
\end{table*}

The dataset is organized by geographical region (africa, france, india, spain). Within each region, subfolders correspond to tile identifiers as shown in Figure~\ref{fig:study_regions}.

\section*{Experimental design, materials and methods}

The dataset was generated through a multi-step pipeline combining satellite data preprocessing, radiative transfer model inversion, and forward hyperspectral simulation. Sentinel-2 Level-2A (\cite{sentinelhub2023}) products were first selected and harmonized to obtain spatially consistent multispectral inputs across four geographic regions. These inputs were then inverted using a PROSAIL-based look-up table to estimate pixel-level vegetation biophysical parameters. Finally, the retrieved parameters were used to drive a forward PROSAIL simulation to produce hyperspectral reflectance cubes. The following subsections describe each component of this pipeline in detail.

\subsection*{Sentinel-2: Data Preprocessing and Image Selection}

\begin{figure*}[!h]
    \centering
    \includegraphics[width=\textwidth]{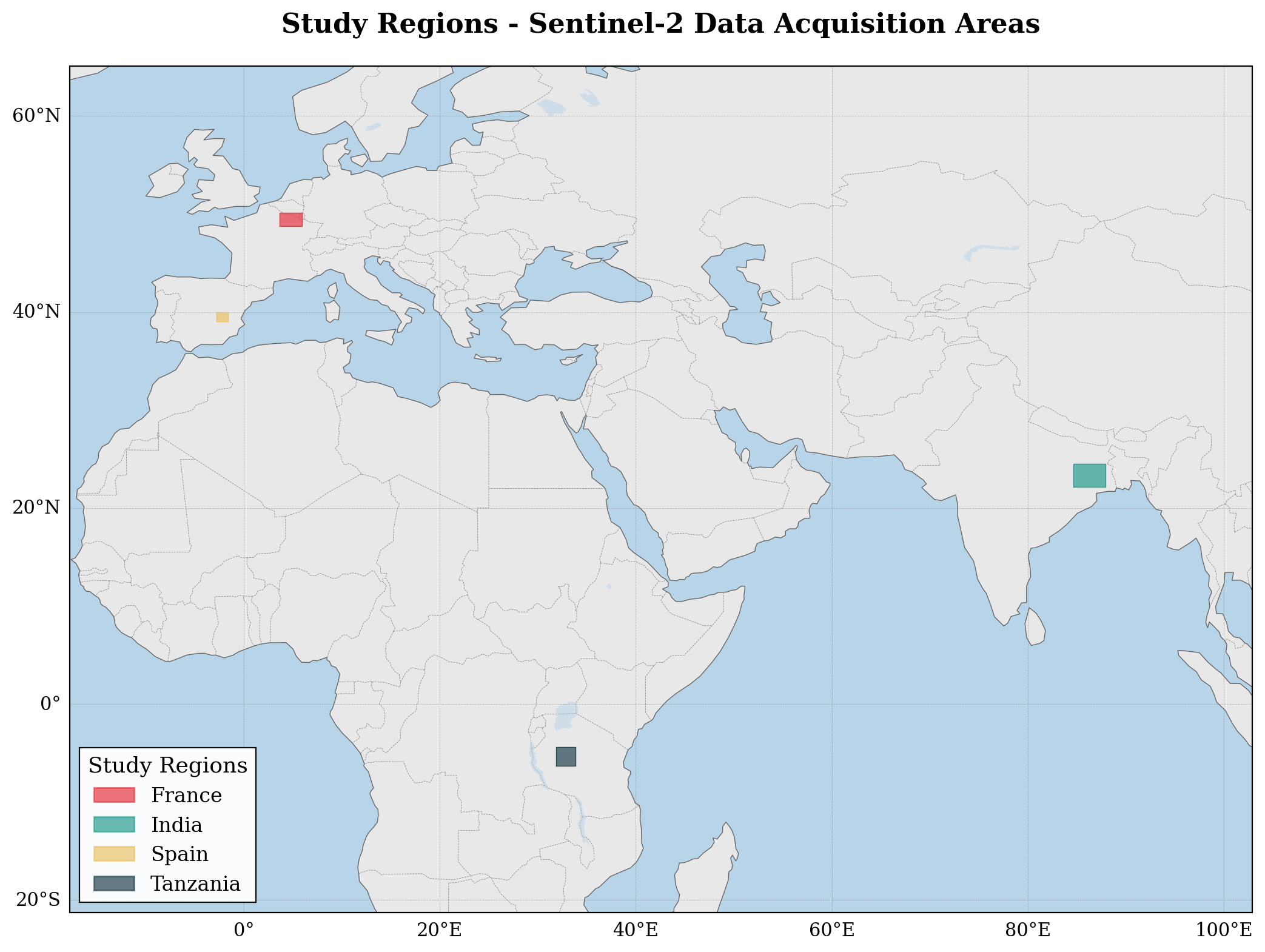}
    \caption{Geographic distribution of the four study regions used for synthetic hyperspectral dataset generation. The regions span diverse climate zones and vegetation types across Europe (France, Spain), Africa (Tanzania), and Asia (India).}
    \label{fig:study_regions}
\end{figure*}

Sentinel-2 is a multispectral Earth observation mission operated by the European Space Agency (ESA) within the Copernicus Programme. It consists of twin satellites (S2A and S2B) equipped with the MultiSpectral Instrument (MSI), which provides imagery across 13 spectral bands spanning the visible to shortwave infrared regions. These data are widely used for vegetation monitoring, land cover mapping, and retrieval of biophysical and biochemical parameters. The MSI acquires data at three spatial resolutions—10 m, 20 m, and 60 m—depending on the spectral band configuration. The Sentinel-2 tiles used for this dataset were acquired from Level-2A surface reflectance products (S2MSI2A) via the Harmonised Data Access service. Four regions—East Africa (Tanzania), Northern France, Spain, and Eastern India—were selected to provide diverse vegetation types and environmental conditions. Scene selection was performed using spatio-temporal queries specifying geographic bounding boxes (Figure~\ref{fig:study_regions} and acquisition dates between 1 and 6 May 2023. All scenes corresponded to platform S2A, instrument MSI, and ONLINE data status. After retrieval, a harmonizing preprocessing pipeline was applied to ensure spectral and spatial consistency across all products. The relevant spectral bands were extracted from each Level-2A tile, specifically ['B1', 'B2', 'B3', 'B4', 'B5', 'B6', 'B7', 'B8', 'B8A', 'B9', 'B11', 'B12']. The 20 m tiles - bands 1-8, bands 11 and 12 - were cropped into patches of $64 \times 64$ pixels, while the 60 m tiles - band 9- were initially cropped at $32 \times 32$ pixels and subsequently upsampled to $64 \times 64$ using nearest neighbour interpolation to maintain uniform spatial resolution across all bands. The resulting 10,883 multispectral cubes are inverted in the next step in order to obtain leaf and canopy biophysical parameters using PROSAIL.

\subsection*{Radiative Transfer Modelling and Look-Up Table Inversion}

The dataset was generated using a two-stage radiative transfer modeling (RTM) pipeline designed to simulate hyperspectral canopy reflectance spectra from vegetation biophysical parameters. This framework integrates two physically based models —PROSPECT-D (\cite{jacquemoud1990prospect,jacquemoud2009}) and SAIL (\cite{verhoef1984light})— which model light interactions at the leaf and canopy scales, respectively.

\paragraph{Leaf-Level Simulation: PROSPECT-D}

The PROSPECT-D model simulates leaf reflectance and transmittance as a function of biochemical and structural leaf properties. The model requires six input parameters: the leaf structure parameter $N$ (number of effective layers), chlorophyll a+b content $C_{ab}$ ($\mu g/cm^2$), carotenoid content $C_{ar}$ ($\mu g/cm^2$), anthocyanin content $C_{ant}$ ($\mu g/cm^2$), brown pigment content $C_{brown}$ (dimensionless), equivalent water thickness $C_w$ ($g/cm^2$), and dry matter content $C_m$ ($g/cm^2$). These parameters collectively describe the optical behavior of leaves across the visible to shortwave infrared spectrum.

\paragraph{Canopy-Level Simulation: SAIL}

At the canopy level, the SAIL model integrates the outputs of PROSPECT-D with structural and environmental variables to simulate canopy reflectance. The required canopy parameters include the leaf area index (LAI, $m^2/m^2$), average leaf inclination angle (ALA, degrees), a hotspot parameter controlling directional reflectance, and the fraction of diffuse incoming radiation. Observation geometry (solar zenith, viewing zenith, and relative azimuth angles) is included to capture bidirectional reflectance effects. Finaly, soil spectra are supplied for each selected region under the assumption of within-region homogeneity, drawn from the ICRAF-ISRIC soil spectral database \cite{DVN2021} using the nearest available measurement site to the target area. The model also computes auxiliary variables such as the fraction of absorbed photosynthetically active radiation ($\textit{f}_{APAR}$) and albedo, which are key indicators of canopy-level energy exchange.

\paragraph{RTM inversion}

To generate the synthetic hyperspectral dataset, Sentinel-2 surface reflectance images were inverted using a lookup-table (LUT) derived from the PROSAIL model to estimate spatially coherent maps of vegetation biophysical parameters. The LUT contained $M = 50\ 000$ simulated spectra generated by sampling the parameter space defined in Table~\ref{tab:prosail_params}. Parameter values were drawn using Latin Hypercube Sampling (LHS) to ensure a homogeneous and statistically efficient exploration of the multidimensional space. For a parameter $\theta_j$ defined over $[\theta_j^{\min}, \theta_j^{\max}]$, the $i$-th LHS sample was computed as
\begin{equation}
\theta_j^{(i)} = \theta_j^{\min} + \frac{\pi_j(i) - U_j^{(i)}}{M} \left( \theta_j^{\max} - \theta_j^{\min} \right),
\end{equation}
where $\pi_j$ is a random permutation of $\{1, 2, ..., M\}$ and $U_j^{(i)} \sim \mathcal{U}(0, 1)$ is a uniform random variable.

Physiological constraints were applied during LUT construction following \cite{danner2021}, notably the empirical coupling between chlorophyll content ($C_{ab}$) and leaf area index (LAI), ensuring that sampled parameters corresponded to realistic vegetation states. Additional plausibility checks were performed at the spectral level: simulated spectra exhibiting green peaks at wavelengths lower than 547\,nm were removed, following the criterion of \cite{wocher2020}. Each retained LUT entry consisted of a parameter vector $\boldsymbol{\theta}^{(i)}$ and its corresponding PROSAIL-simulated reflectance spectrum $\rho^{(i)}(\lambda)$. The inversion step consisted of matching observed Sentinel-2 reflectance $\rho^{\mathrm{obs}}$ to the simulated LUT entries. For each pixel, spectral discrepancy was quantified using the root mean square error (RMSE) as a cost function,
\begin{equation}
\mathrm{RMSE} = \sqrt{\frac{1}{211} \sum_{b=1}^{211} \left( \rho^{\mathrm{obs}}_b - \rho^{\mathrm{sim}}_b \right)^2}.
\end{equation}
For each pixel, $n=10$ LUT entries with the lowest cost values were retained, forming an ensemble of plausible biophysical solutions. Final parameter estimates were derived using the median. Parameter uncertainty was assessed non-parametrically from the 5th and 95th percentiles of the $n$-best ensemble, providing pixel-level confidence intervals. The resulting spatially explicit maps of PROSAIL biophysical parameters were then used as inputs to a forward PROSAIL simulation chain to generate pixel-level hyperspectral reflectance spectra. Each simulated spectrum represents canopy-scale reflectance under the geometric, physiological, and structural conditions estimated during the inversion.

\subsection*{Visualisation}

\begin{figure*}[htb]
    \centering
    \includegraphics[width=\textwidth]{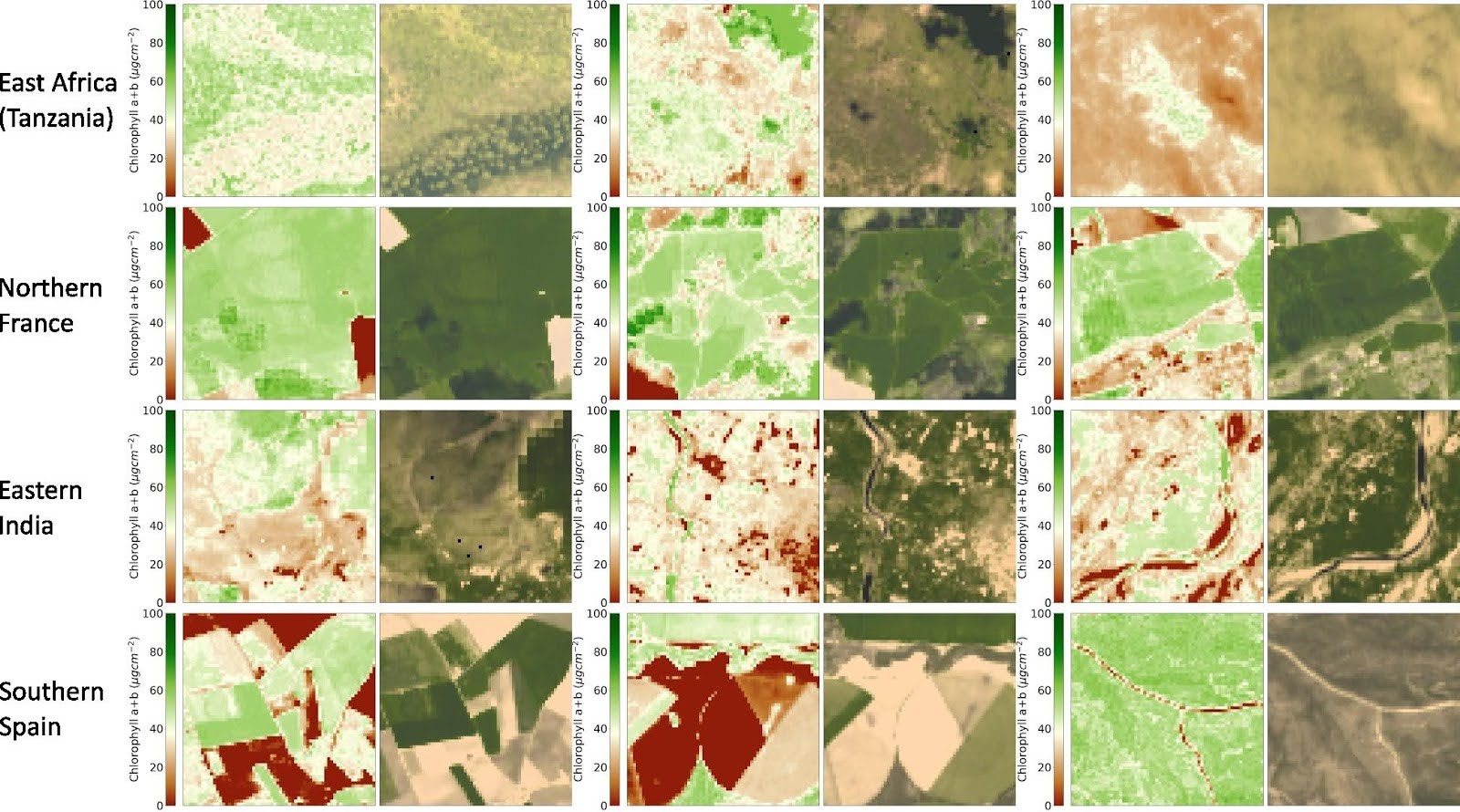}
    \caption{Representative examples of emulated hyperspectral image cubes and their corresponding chlorophyll a+b (Cab) maps from the four study regions. For each row, the left column shows the Cab map and the right column shows the associated HSI scene. All scenes are displayed using an RGB composite.}
    \label{fig:scenes}
\end{figure*}

\begin{figure*}[!ht]
    \centering
    \begin{tabular}{cccc}
        \includegraphics[width=0.23\textwidth]{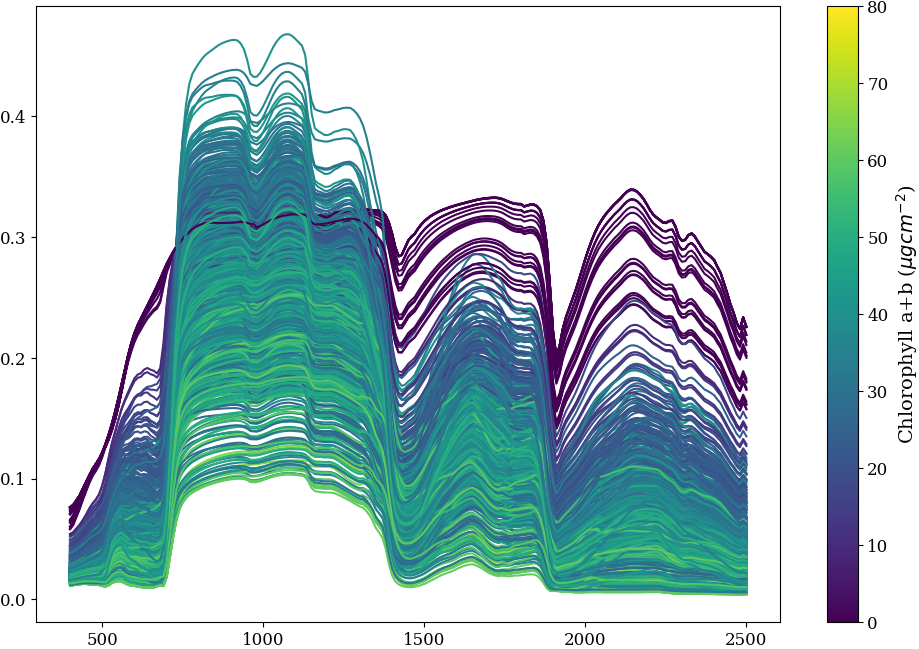} &
        \includegraphics[width=0.23\textwidth]{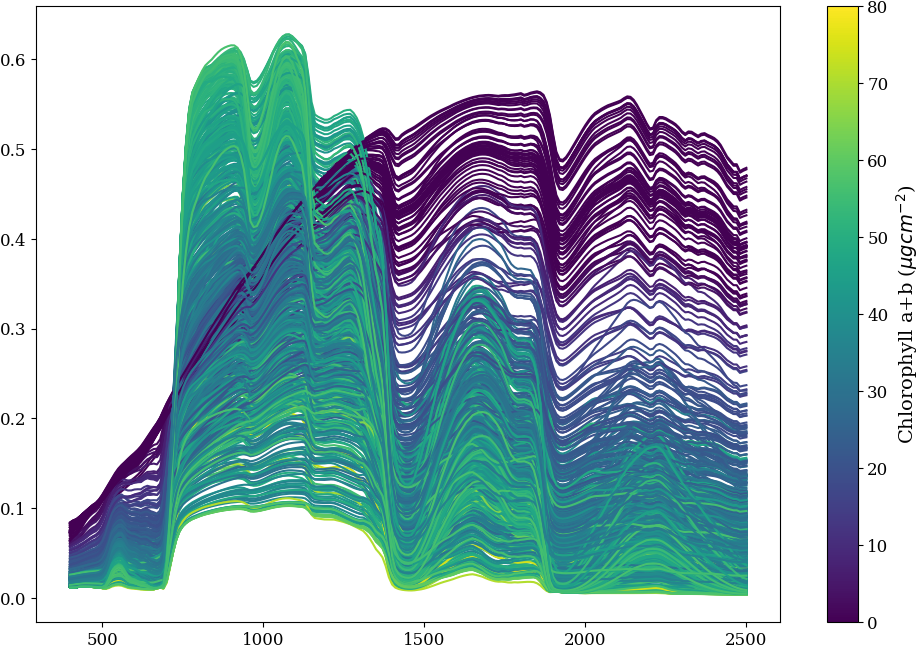} & 
        \includegraphics[width=0.23\textwidth]{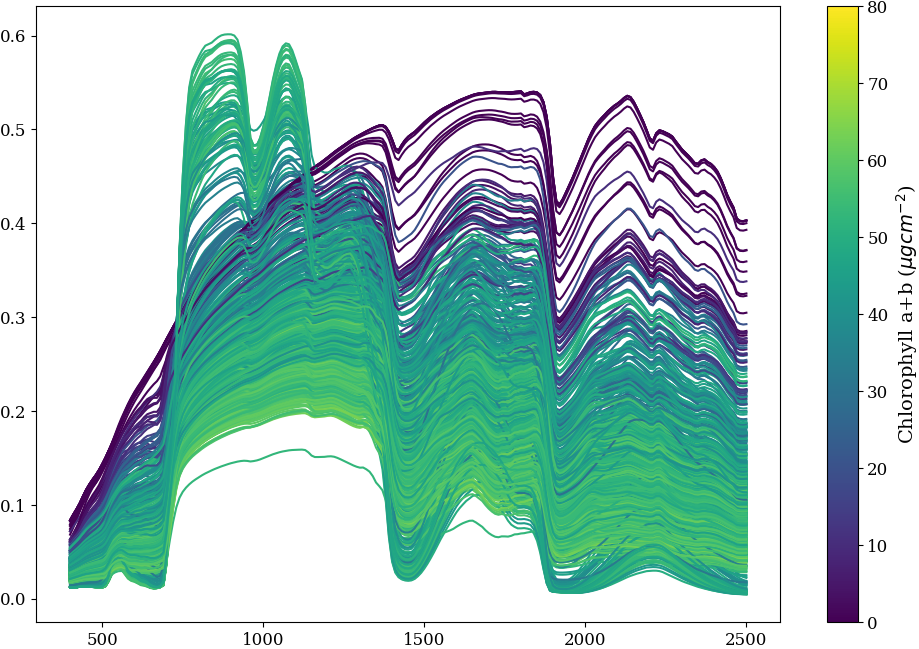} &
        \includegraphics[width=0.23\textwidth]{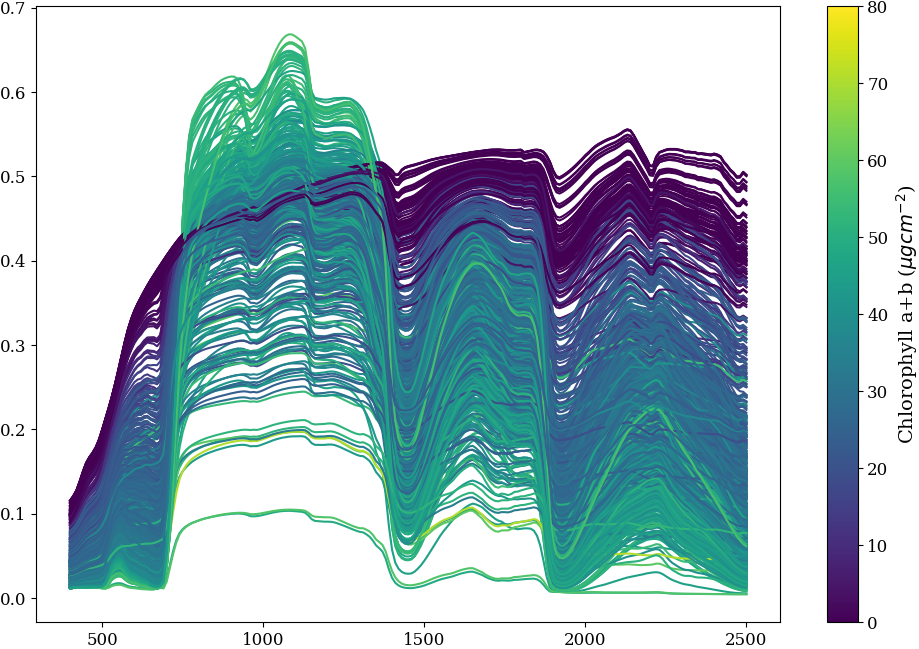} \\
        
        Tanzania (East Africa) & Northern France & Southern Spain & Eastern India \\
        
    \end{tabular}
    
    \caption{Simulated hyperspectral reflectance spectra from the four study regions, color-coded by chlorophyll a+b content ($C_{ab}$). }
    \label{fig:regional_spectra}
\end{figure*}

Representative examples of HSI cubes from each region are shown in Figure~\ref{fig:scenes}. These samples illustrate the visual diversity of the dataset in terms of canopy cover, landscape structure, and spectral signatures. For each region, the figure displays the chlorophyll a+b (Cab) map alongside its corresponding RGB composite HSI scene, allowing simultaneous inspection of biochemical variability and spatial reflectance patterns. The examples highlight transitions between cropland parcels, forested clusters, semi-arid shrublands, and heterogeneous mosaics characteristic of the selected environments. In addition, the spectral behavior associated with these land-cover configurations is illustrated in Figure~\ref{fig:regional_spectra}, which shows representative reflectance spectra for each region, color-coded by Cab concentration to emphasize the relationship between vegetation biochemistry and spectral shape.

\section*{Limitations}

While the dataset covers four ecological regions, it does not capture the full diversity of global vegetation or environmental conditions. Additionally, each region is represented using a single soil type, which does not reflect soil variability and therefore the full range of soil‑driven effects on bio‑optical parameters. Finally, the synthetic generation process also relies on strong assumptions in both the forward and inverse modeling steps, which should be considered when applying the dataset beyond controlled experimental settings.

\section*{Ethics statement}

The authors have read and follow the ethical requirements for publication in Data in Brief. The current work does not involve human subjects, animal experiments, or any data collected from social media platforms.

\section*{CRediT author statement}

\textbf{Chedly Ben Azizi}: Conceptualization, Methodology, Software, Validation, Data Curation, Writing – Original Draft.\textbf{ Claire Guilloteau}: Conceptualization, Validation, Writing – Reviews \& Editing, Supervision, Funding acquisition.\textbf{ Gilles Roussel}: Conceptualization, Validation, Writing – Reviews \& Editing, Supervision, Funding acquisition. \textbf{Matthieu Puigt}: Writing – Reviews \& Editing.

 \section*{Declaration of competing interests}

 The authors declare that they have no known competing financial interests or personal relationships that could have appeared to influence the work reported in this paper.

\section*{Acknowledgments}
 This work is partially funded by the ULCO research pole ''Mutations Technologiques et Environnementales'', EUR MAIA (ANR-22-EXES-0009) and Hauts-de-France region. Experiments presented in this paper were carried out using the CALCULCO computing platform, supported by DSI/ULCO (Direction des Systèmes d'Information de l’Université du Littoral Côte d’Opale).

\bibliographystyle{IEEEtran}
\bibliography{IEEEabrv,refs}

@article{jacquemoud1990prospect,
  title={PROSPECT: A model of leaf optical properties spectra},
  author={Jacquemoud, St{\'e}phane and Baret, Fr{\'e}d{\'e}ric},
  journal={Remote sensing of environment},
  volume={34},
  number={2},
  pages={75--91},
  year={1990},
  publisher={Elsevier}
}

@article{verhoef1984light,
  title={Light scattering by leaf layers with application to canopy reflectance modeling: The SAIL model},
  author={Verhoef, Wouter},
  journal={Remote sensing of environment},
  volume={16},
  number={2},
  pages={125--141},
  year={1984},
  publisher={Elsevier}
}

@article{jacquemoud2009,
title = {PROSPECT+SAIL models: A review of use for vegetation characterization},
author = {Stéphane Jacquemoud and Wout Verhoef and Frédéric Baret and Cédric Bacour and Pablo J. Zarco-Tejada and Gregory P. Asner and Christophe François and Susan L. Ustin},
journal = {Remote Sensing of Environment},
volume = {113},
pages = {S56-S66},
year = {2009},
note = {Imaging Spectroscopy Special Issue},
issn = {0034-4257},
doi = {https://doi.org/10.1016/j.rse.2008.01.026}
}

@misc{DVN2021,
author = {World Agroforestry (ICRAF) and International Soil Reference and Information Centre (ISRIC)},
publisher = {World Agroforestry (ICRAF)},
title = {{ICRAF-ISRIC Soil VNIR Spectral Library}},
UNF = {UNF:6:gDRzU59a0YAKsRM4nQDdPg==},
year = {2021},
version = {V1},
doi = {10.34725/DVN/MFHA9C},
url = {https://doi.org/10.34725/DVN/MFHA9C},
note = {Dataset}
}

@online{sentinelhub2023,
  author = {{European Space Agency}},
  title  = {Modified Copernicus Sentinel data 2023},
  year   = {2023},
  note   = {Accessed via Sentinel Hub},
  url    = {https://www.sentinel-hub.com/}
}

@article{danner2021,
  title={Efficient RTM-based training of machine learning regression algorithms to quantify biophysical \& biochemical traits of agricultural crops},
  author={Danner, Martin and Berger, Katja and Wocher, Matthias and Mauser, Wolfram and Hank, Tobias},
  journal={ISPRS Journal of Photogrammetry and Remote sensing},
  volume={173},
  pages={278--296},
  year={2021},
  publisher={Elsevier}
}

@article{wocher2020,
  title={RTM-based dynamic absorption integrals for the retrieval of biochemical vegetation traits},
  author={Wocher, Matthias and Berger, Katja and Danner, Martin and Mauser, Wolfram and Hank, Tobias},
  journal={International Journal of Applied Earth Observation and Geoinformation},
  volume={93},
  pages={102219},
  year={2020},
  publisher={Elsevier}
}

\end{document}